\documentclass[conference]{IEEEtran}
\IEEEoverridecommandlockouts
\usepackage{amsmath}
\usepackage[caption=false]{subfig}
\usepackage{tabularx,graphicx}
\usepackage{hyperref}
\usepackage[numbers,sort&compress]{natbib}
\usepackage{amsfonts}
\usepackage{tabularx,multirow}
\usepackage{adjustbox}

\begin{document}

\title{Behavioral Player Rating in Competitive\\ Online Shooter Games}

\author{\IEEEauthorblockN{Arman Dehpanah}
\IEEEauthorblockA{\textit{School of Computing} \\
\textit{DePaul University}\\
Chicago, USA \\
\small{adehpana@depaul.edu}}
\and
\IEEEauthorblockN{Muheeb Faizan Ghori}
\IEEEauthorblockA{\textit{School of Computing} \\
\textit{DePaul University}\\
Chicago, USA \\
\small{mghori2@depaul.edu}}
\and
\IEEEauthorblockN{Jonathan Gemmell}
\IEEEauthorblockA{\textit{School of Computing} \\
\textit{DePaul University}\\
Chicago, USA \\
\small{jgemmell@cdm.depaul.edu}}
\and
\IEEEauthorblockN{Bamshad Mobasher}
\IEEEauthorblockA{\textit{School of Computing} \\
\textit{DePaul University}\\
Chicago, USA \\
\small{mobasher@cs.depaul.edu}}
}


\maketitle

\begin{abstract}
    Competitive online games use rating systems for matchmaking; progression-based algorithms that estimate the skill level of players with interpretable ratings in terms of the outcome of the games they played.
    However, the overall experience of players is shaped by factors beyond the sole outcome of their games.
    In this paper, we engineer several features from in-game statistics to model players and create ratings that accurately represent their behavior and true performance level.
    We then compare the estimating power of our behavioral ratings against ratings created with three mainstream rating systems by predicting rank of players in four popular game modes from the competitive shooter genre.
    Our results show that the behavioral ratings present more accurate performance estimations while maintaining the interpretability of the created representations.
    Considering different aspects of the playing behavior of players and using behavioral ratings for matchmaking can lead to match-ups that are more aligned with players' goals and interests, consequently resulting in a more enjoyable gaming experience.
\end{abstract}


\section{Introduction}

Competitive online games have revolutionized gaming experience across different gaming platforms.
In games such as Fortnite, you can choose to play as \emph{Kratus}, the well-known protagonist from the popular ``God of War'' game series, and fight against another player who happens to be \emph{Spider-man}.
These games have been so well-received that publishers of offline games have actively started to develop online versions of their games to enter this ever-increasing market.
These games use matchmaking systems to create balanced match-ups and ensure an enjoyable experience for players.
 
Matchmaking systems use previous performance of players to model their current level of performance.
These systems then use the created models to assign players to teams and match-up teams for different events.
Matchmaking systems differ in the way they model players.

Rating systems are a specific type of matchmaking systems that use statistical inference to model the performance of players from the outcome of their previous games.
Since the only data these systems use is the outcome of games, they are fast and interpretable. 
Other matchmaking systems use machine learning techniques to predict the performance of players from a wide range of in-game statistics and their corresponding latent factors.
Due to using extra information, the player models created by these algorithms are often a better representative of the actual performance of players. 
 
While both of these systems are widely used in games, they are often either too simple or too complex.
Looking solely at the outcome of the games may lead to losing potentially useful information existing in the gameplay data.
On the other hand, using an excessive amount of gameplay data and creating latent features based on those data may lead to intensive memory consumption and inexplicable player models.

This study aims to incorporate the strengths of both of these approaches while alleviating their shortcomings.
To that end, we conduct extensive feature engineering on several real-world datasets of popular game modes from the competitive shooter genre to capture different aspects of the behavior of players.
We then build our player/team models around the engineered behavioral features to create behavioral ratings.
The created ratings carry the whole playing history of players and all the meaningful interactions they had within the games they played.
These ratings are updated after each game based on the events occurred within the game.

We hypothesize that our behavioral ratings accurately capture the true performance levels of players while maintaining the interpretability of their representations.
To test our hypothesis, we use the most recent behavioral ratings to predict the performance of players and teams (i.e., their ranks) for their upcoming matches.
The prediction results are then compared to those of three popular rating systems; Elo, Glicko, and TrueSkill.
Our results show that behavioral ratings outperform popular rating systems in rank prediction, specially when combined using optimal weights.


\section{Related Work}
\label{sec:related}

Competitive online games such as Apex Legends and Call of Duty: Warzone serve millions of players on a daily basis.
Due to the abundant playing options for players, to stay alive, these games need to maintain and expand their player base.
A core principle to accomplish this task is to provide players with a pleasant experience to keep them engaged with the game.
To ensure such a pleasant experience, competitive online games use matchmaking systems.

Matchmaking systems often differ in objectives they pursue for conducting assignments.
While some systems focus on player engagement~\cite{chen2017eomm} and enjoyment~\cite{delalleau2012beyond}, the majority of matchmaking algorithms aim for maintaining competitive balance in matches they create~\cite{nikolakaki2020competitive}.
These algorithms assume that the most pleasant experience happens when the match is optimally balanced, meaning the odds of winning are more or less even for the competing players and teams~\cite{geertz2000deep}.
Therefore, in most cases, a key step of their assignment process includes predicting the probability of winning and consequently, estimating the potential performance for possible combination of players and teams.

Rating systems model players' performance based on the outcome of games they played in the past.
These systems use numerical estimates, referred to as ratings, to model a player's skill level.
Skill ratings are used to calculate the probability of winning for each side of a match.
The calculated probabilities are then used to predict rank and create match-ups.

Rating systems are extensively used for matchmaking and creating leaderboards in many games.
These systems are very popular due to the interpretability and intelligibility of the ratings they create.
The ratings also provide players with a tool for following their skill development. 
Despite their widespread use, these systems, in their traditional form, model players purely based on their skill level inferred from the outcome of matches they played, and thus, lose potentially useful information lied in the gameplay data.

Modern games often include complex structures where players need to master multiple aspects of the gameplay to improve their performance.
Such aspects may range from learning how to properly use a weapon or special power of a character to how to communicate with teammates and how to strategize.
We refer to these aspects as the behavior of players.
While rating systems define the behavior of players solely based on their skill level, it is reasonable to assume that many factors other than skill play a role in determining the overall behavior of players.
Such factors are often recorded in gameplay data and thus, can be derived from previous performance of players in the matches they played. 
Efforts have been made to alleviate this issue by extending upon traditional rating systems using different aspects of gameplay such as context~\cite{chen2016predicting}, chemistry between team members~\cite{delong2011teamskill}, offensive/defensive skills~\cite{stanescu2011rating}, victory and draw margin~\cite{kovalchik2020extension, dangauthier2007trueskill}, experience~\cite{minka2018trueskill}, and time~\cite{coulom2008whole}.

While features such as offensive skills or experience are well-known to a typical audience, others may be of a more tacit nature or come from a combination of other features and thus, may not have a clear definition.
To capture relevant information determining the behavior of players from both of these feature groups, research often leverages predictive models that work based on generating latent factors from explicit historic data~\cite{delalleau2012beyond,chen2016predicting}.
These models typically receive known behavioral features as input and then build latent features by weighting, combining or transforming those inputs.
The outcome of these models are a prediction score signifying a utility such as the match balance.
Such models often achieve better prediction results and higher quality matchmakings compared to rating systems. However, they lack explainability due to the complex nature of their calculations; an important feature for players who are interested in understanding the logic behind their match-ups~\cite{seif2020data}.

In this paper, we focus on eliminating the limitations of both latent factor models and rating systems while maintaining their strengths.
To this end, we conduct extensive experimentation on several real world datasets to engineer behavioral features from raw game statistics.
The engineered features represent different aspects of the playing behavior of players.
We use those behavioral features to create interpretable player and team ratings.
We then predict the performance of players and teams (i.e., their ranks) and compare the prediction results with those of three popular rating systems; Elo~\cite{elo1978rating}, Glicko~\cite{glickman1999parameter}, and TrueSkill~\cite{herbrich2007trueskill}.

Each game genre is often associated with certain playing behaviors.
For example, strategy games require intricate planning, intelligent tactics, and significant collaborations among teammates while shooter games call for expeditious decision-making and meticulous shooting.
While the majority of previous efforts focused on a single game mode, or multiple games from different genres, in this paper, we specifically focus on competitive shooter games and consider several game modes within this genre to achieve a rigorous understanding of players' behavior in these games.
Finally, in contrast with complex mathematical computations and statistical estimations used in previous works, we engineer behavioral features based on players' previous performance and use those features to create simple and interpretable behavioral ratings that represent different aspects of a player's behavior.
Such ratings provide an accurate representation of players' performance while being easily explainable and interpretable to a non-technical audience.


\section{Behavioral Modeling and Game Modes}
\label{sec:modeling}

In this section, we introduce a number of popular game modes in competitive shooter games and discuss gameplay features that can potentially represent different aspects of a player's behavior within those modes.
We then explain how such features can be used to model players and teams.

\subsection{Game Modes}
\label{sec:gamemodes}

In general, competitive shooter games can be divided into two groups in terms of the number of the sides competing against each other.
The first group is referred to as head-to-head games where two players or teams face each other and engage in fights.
This group consists of either `one-versus-one' games such as one on one duels in Quake Champions or `n-versus-n' games such as a typical competitive 5 on 5 game in Counter Strike: Global Offensive (CS:GO).
As opposed to the first group, games in the second group, referred to as free-for-all games, involve more than two players or teams simultaneously competing in the same match.
These games are either played as `solo' where many individual players compete against each other as singletons, or as `squad' (e.g., duos, trios, or quads) where multiple teams of more than one player face each other in the same match.

Competitive shooter games also differ in the way the winner of a match is decided.
In some games, players win the match if they achieve the highest score at the end while in others, players who survive the most will win the match.

In this paper, we consider four popular modes; Deathmatch, Capture the Flag, Tactical Head-to-Head, and Battle Royale.

\subsubsection{Deathmatch}
Deathmatch is one of the most popular game modes extensively played in games such as Halo, Quake, Doom, and CS:GO.
A deathmatch game can be played as either head-to-head or free-for-all.
In this mode, players or teams engage in direct fights and obtain scores upon eliminating an opponent.
The key characteristic of deathmatch is that players who are eliminated keep respawning endlessly until the end of the match.
The winner of a deathmatch game is often decided by setting a time or score threshold.
With a time threshold, the match ends upon reaching a time limit and the player or team with the highest score is the winner of the match.
With a score threshold, in addition to the above rule, the player or team who first reaches a pre-determined score wins the match.

Compared to most game modes, deathmatch games are often very fast-paced with relatively small maps (i.e., the area in which the game is running) which causes a myriad of quick encounters and chaotic fights. 
This is specifically true in casual games.
On the other hand, in competitive ranked games, teams often incorporate limited strategies in their play.
However, due to the nature of the fights, committing to such strategies becomes impossible at times, and thus, individual performance is much more highlighted in this game mode.

In addition, as opposed to most game modes, in deathmatch there is often no restriction on the weapons players can carry.
Various types of weapons can be selected upon (re)spawn or be picked up from different locations all around the map.
While some weapons are more effective than others, the type of weapon does not change the scoring function; players obtain one point for every enemy they eliminate regardless of the weapon they used.

Considering all of the above aspects, in deathmatch mode, features that capture players' aggressiveness, expertise, skills, and experience may better represent their behavior compared to features that imply their strategies.

\subsubsection{Capture the Flag (CTF)}
CTF is similar to a head-to-head team-based deathmatch in that players respawn indefinitely upon elimination.
Also, in CTF players typically start the game with pre-selected weapons but like deathmatch, can pick up different weapons and items scattered all around the map.
CTF maps are relatively larger than deathmatch maps but similar to deathmatch, CTF games are fast-paced filled with chaotic direct fights.
Therefore, mechanical skills of individual team members play an important role in success of teams.
However, due to relatively larger playing area, teams can implement various tactics and strategies when defending or attacking an area.
The main difference between the two modes is that unlike deathmatch, CTF is an objective-based mode.

In CTF, although eliminating opponents increases the chance of winning, it does not necessarily lead to victory.
Each competing team in CTF starts the game in their assigned bases with a flag.
The teams have to defend their own flag while trying to capture the opponent's flag.
When a player steals the opponent's flag, they become the `flag bearer'.
A flag bearer has to carry the flag to their own base to capture the flag and earn a point, provided that their own flag is not in transit.
The game often imposes a time limit after which the team with higher points wins the match.

Teamwork plays a crucial role in success of teams in a CTF match.
In order to succeed, a team has to assign some players to defend the flag in its base while sending others to attack the enemy's base and steal their flag.
These roles vary entirely depending on the ongoing events.
For example, when a team's flag is stolen, the majority of its players attempt to attack the flag bearer opponent to eliminate them and retrieve the flag.
On the other hand, the teammates of a flag bearer put their whole focus on defending the stolen flag particularly because the flag bearer cannot use weapons while carrying the flag.

Finally, due to relatively large maps and imposing a time limit, CTF often includes modes of transport such as vehicles and cannons to facilitate faster travel between bases.
Vehicles in particular can be helpful when capturing a flag as the flag bearer can board a vehicle run by a teammate to avoid attacks and reach the base faster to capture the flag.
On the other hand, a vehicle is a bigger target and easier to hit by enemy's attacks.
Therefore, players who are experienced in driving in-game vehicles are usually assigned to this task.

Considering all of the above aspects, in CTF mode, features that capture players' expertise, skills, and experience along with features that imply their tactical abilities may sufficiently represent their playing behavior.

\subsubsection{Tactical Head-to-Head}
Tactical head-to-head shooters are one of the most traditional games in the realm of competitive games.
Call of Duty: Modern Warfare, CS:GO, and Valorant are among the most popular games offering this game mode where two teams consisting of equal number of players compete against each other in a head-to-head fight.
What the majority of games in this mode have in common is an objective-based gameplay.
Two teams are divided into attacking and defending sides pursuing objectives such as detonating/defusing a bomb or capturing/rescuing hostages.
Also, the maps in this mode include strategically important locations and are relatively larger compared to the maps used in deathmatch games to accommodate both random fast-paced and tactical plays.

Another common element of games in this mode is an in-game currency system.
Players receive points in the form of a currency as they complete objectives or eliminate opponents within a match.
Such currencies can then be used to purchase desired weapons and items.
As such, being well-acquainted with the mechanisms of the buy menu and items offered in the system is extremely important.

In addition to the above elements, perhaps the most important common aspect in tactical head-to-head shooters distinguishing this mode from other modes in the competitive shooter genre is the necessity of having a perfect harmony between individual players' expertise and team strategies and tactics to succeed.
Mechanical skills such as good aiming still play an important role in this mode, but relying solely on such skills without adopting efficient tactics and strategies for the team often cost too much compared to other modes.

Considering all of the above aspects, in tactical head-to-head mode, features that capture both mechanical and tactical aspects of players may better represent their behavior.

\subsubsection{Battle Royale}

Battle royale is a free-for-all game where many players and teams compete against each other within the same match at the same time.
In this mode, players parachute into the playing area from a plane with minimal equipment and quickly start looting the area to find weapons and items.
The goal is to be the last player or team standing to win the match.
Players can either directly engage in fights from the start to eliminate everyone else and win the game or choose to `camp' and avoid fights to increase their chances of survival.

Battle royale maps are typically larger compared to those of other modes due to higher number of players competing in this mode.
For example, a typical battle royale match in Call of Duty: Warzone consists of 150 players as opposed to a typical head-to-head CS:GO match with only 10 players.
To enable players to travel such large spaces faster, battle royale games often comprise of different modes of transportation such as vehicles, jump towers, portals, or cannons.
These items can be used by players for different purposes from finding opponents faster to escaping from a losing battle.
As opposed to other modes, battle royale does not impose a time limit.
Instead, the games in this mode include a constantly shrinking safe area on the map outside of which players take damage.
This way, players are forced to engage in fights.

While weapons and items are randomly scattered around the map, in some locations, referred to as `hot zones', players have a higher chance of finding more powerful equipment.
The first fights of the match usually happens in these areas as many players land in hot zones with the hope of finding better equipment.
In such cases, knowing key places within those areas and having quick reflexes for faster looting play an important role in winning early fights.

Due to the large playing area, battle royale games are not as fast-paced as deathmatch games.
This allows players and teams to adopt appropriate strategies and tactics in their fights.
However, since the game includes many players and teams, there is a good chance that such tactics and strategies are interrupted by a `third-party'; another player or team who seeks to attack two sides during their ongoing fight.

Considering all of the above aspects along with the fact that battle royale games include smaller teams compared to other modes, individual skills and team performance are both crucial in determining the success of a team.
Therefore, features that capture both mechanical and tactical aspects of players may better represent their behavior in the battle royale mode.

\subsection{Capturing Behavioral Features}
In the previous section, we mentioned a number of features such as aggressiveness, experience, expertise, and tactical abilities that represent a player's behavior.
In this section, we describe specific in-game statistics that could demonstrate those features.

A common goal players often pursue in competitive shooter games is eliminating their opponents.
As mentioned earlier, opponent elimination is a deciding factor in winning the game in some modes while in others it may be rather beneficial and sometimes even a futile attempt.
Regardless, looking at the number of opponents eliminated by a player, or more commonly referred to as `number of kills' in shooter games, may represent how much they favor aggressive play.
However, looking solely at the number of kills may be misleading.
For example, a player who scored 50 kills within 10 games can be considered aggressive and competitive but a player who scored the same number of kills in 100 games is perhaps a casual player.
A more reliable approach is to consider the number of kills against the number of times the player was eliminated.
A popular metric conforming to this requirement is the ratio between the number of kills to the number of times the player was eliminated, otherwise known as `Kill to Death or KD Ratio'.
Players with higher KD ratio are assumed to be more aggressive in their playing style.
In competitive shooter games, being aggressive is synonymous with being highly skilled as this playing style is often observed in top-tier players while amateur and low-tier players frequently adopt a passive playing style.
Looking at the outcome of matches, like traditional rating systems do, can also help capture these behaviors.
Players who frequently win their matches can be considered high-skilled or top-tier players while players who constantly lose can be labeled as amateur or low-tier players.

Expertise is another important behavioral factor affecting the performance of players.
Expertise may include various components.
For example, some players may be an expert at shooting a weapon while others are expert in using grenades or melee attacks; some players may be die-hard survivors while others may be adept drivers.
Such components can be taken into account by looking at different in-game statistics.
Being expert at shooting a weapon may be represented by accuracy of shooting.
In shooter games, being accurate means to shoot an area that inflicts the highest amount of damage, and that is usually the opponent's head.
Therefore, the average number of headshots a player registered in the games they played can be a proper indication of their shooting expertise.
Also, the average number of kills scored using grenades or melee attacks can represent their expertise for each case.
In addition, statistics such as the average amount of time a player could stay alive during the matches they played can represent their survival skills.
Similarly, statistics such as the average number of times a player used in-game vehicles or the average distance they traveled using such vehicles may indicate their expertise and interest in driving.
Average distance traveled on foot, on the other hand, is an important measure for capturing both playing style and tactics.
Regardless of game mode, players who traveled a greater distance on foot in matches they played can generally be considered aggressive players since this type of players are consistently in motion searching for opponents to eliminate.
On the contrary, smaller values for distance walked may suggest a `camper'; a player who chooses a safe spot to hide and waits for ambushing a passerby opponent.
However, in tactical head-to-head or CTF games where teams adopt various strategies and tactics, players who walked smaller distances may be the ones assigned to defending strategic locations such as a bomb site or flag base.

Finally, the experience level of players is a crucial measure circumscribing all behavioral factors.
While each game mode has its own learning process, as a general rule, it is assumed that the more players play a game, the more experience they gain in that game.
Therefore, the number of games played can be considered as a fitting manifestation of a player's experience level also showing how much effort players put into the game and how engaged they are with the game.

\subsection{Behavioral Modeling and Rating}
We engineer behavioral features based on available statistics recorded in the data to create player models.
Such player models are represented by corresponding behavioral feature values denoted by $\mu$ and referred to as single-factor models.
All features are normalized using Z-score to achieve consistency and avoid scaling issues.
Similar to skill ratings created by rating systems, these values can be considered as ratings that offer a basis for comparing the performance of players from different behavioral aspects. 

We also calculate a combination of behavioral features to create inclusive behavioral ratings that cover different aspects of a player's performance.
We aggregate features using a linear combination to ensure the calculated ratings are explainable.
We consider a non-weighted linear hybrid of features as our naive inclusive behavioral rating, denoted by $\eta$, that is a simple summation of all normalized feature values.
In this approach, having $n$ behavioral features, the behavioral rating of player $p$ is calculated as:
\begin{equation*}
    \eta(p) =  \sum^{n}_{i=1}\mu_i(p)
\end{equation*}

\noindent where $\mu$ represents normalized individual feature values.
The naive hybrid approach takes into account all engineered behavioral features, regardless of their relevance to or impact on the overall player's performance.

When engineering behavioral features from in-game statistics, a key aspect worth exploring is detecting a higher level meaning that a group of behavioral features may imply.
Such features are often highly correlated with each other; an undesired characteristic of data that may cause overfitted and unnecessarily complex models.
Factor Analysis~\cite{harman1976modern} is a popular approach for overcoming this issue.
In this method, a group of observed features are decomposed into smaller number of unobserved factors that capture most of the variance accounted for by those features.
These factors are then rotated to result in interpretable representations which are shown as feature loadings for each extracted factor.
Therefore, for player $p$, a factor $\mathcal{F}$ can be considered as a linear combination of its underlying features and their corresponding loadings:
\vspace{-2.mm}

\small
\begin{equation*}
    \mathcal{F}(p) = \sum^{n}_{i=1} \ell_i(p) \: \mu_i(p)
\end{equation*}
\normalsize
\vspace{-1.5mm}

\noindent where $\mu$ denotes a behavioral feature and $\ell$ shows its loading onto factor $\mathcal{F}$.
For example, features such as KD ratio, accuracy, and winning rate may all imply how skilled a player is, and thus, may be aggregated into a factor representing the concept of \emph{skill}.
We perform factor analysis on engineered behavioral features to detect inter-dependencies between features and capture higher-level concepts representing those features.

Each constructed behavioral feature (or factor) may differently contribute to the overall performance of players and consequently, to the performance of their teams.
Some features may represent a positive performance while others may suggest interactions that negatively affect the players' performance and their rank.
For example, high-skilled aggressive players who lead their team effort often contribute to their team's success while players who often tend to play toxic and eliminate their own teammates on purpose hurt both their and their team's chance of winning.
The constructed behavioral ratings should address the nature of behavioral features and capture the direction and magnitude of their effects on the overall performance of players.
To achieve a more accurate representation of ratings, we need to identify the most relevant behavioral features along with their degree of contribution indicated by feature weights.
To this end, logistic regression can be used to classify the outcome of matches with behavioral features as predictors.

A logistic regression, being a generalized linear model, estimates the odds of a given outcome category against other possible categories using a linear weighted combination of observed predictors.
Since the significance of contribution is statistically tested for each predictor in logistic regression, the model can be further improved by discarding irrelevant features using feature selection and regularization.
The final logistic regression model yields a weighted linear combination of important features; an equation that predicts the (log) odds of possible outcomes considering feature values. 
The resulted weights determine the degree to which each feature affects the outcome and thus, can be used to weight behavioral features in constructing more accurate behavioral ratings.

Binary logistic regression can be used for classifying the outcome of head-to-head games since the outcome is either win or loss in our modes of interest.
On the other hand, ordinal logistic regression is a better fit for battle royale games since their outcome is a list of ranks where the order between possible values matters.
Regardless, with $n$ initial behavioral features, the estimated logistic model can be given by:

\vspace{-4.5mm}
\begin{equation*}
    Logit(\hat{Y}) = I + \omega_1\mu_1 + \omega_2\mu_2 + ... + \omega_m\mu_m + \epsilon
\end{equation*}

\noindent where $I$ denotes the intercept, $m$ shows the number of features in the reduced model with $m<n$, and $\omega\mu$ pairs show single-factor ratings and their corresponding weights.
In this equation, features with a positive weight push the prediction toward the class of interest and vice versa.
In an ordinal logistic regression where the dependent variable takes multiple ordered values such as ranks, based on the proportional odds model~\cite{mccullagh1980regression}, the equation remains the same for predicting each rank except for the intercept $I$ that is discarded in our case since we are only interested in relevant behavioral features and their weights to construct more accurate behavioral ratings.
We then normalize the weights to sum up to one to have a more interpretable basis for comparing the effect of features.
When performing this normalization, we consider the absolute value of weights since their signs only show the direction to which the features affect the prediction and not the magnitude of effect itself.
As such, the weighted hybrid behavioral rating for player $p$ is denoted by $\Omega$ and is calculated as:
\begin{equation*}
    \Omega(p) =  \sum^{m}_{i=1}\hat{\omega}_i(p)\mu_i(p)
\end{equation*}

\noindent where $\mu_i$ denotes relevant behavioral features and $\hat{\omega}_i$ shows their corresponding normalized weights.

A machine learning approach like logistic regression by itself can be used for matchmaking by predicting match outcomes~\cite{prasetio2016predicting} without creating any types of player ratings.
However, such approaches do not deliver insights into the different aspects of a player's behavior.
As an alternative, we use logistic regression to weight behavioral features and utilize those weights to create inclusive and accurate behavioral ratings.
Such ratings can be used as a basis for predicting match outcomes and matchmaking while also serving as a tool for comparing player performances and creating leaderboards.

\subsection{Rank Prediction using Ratings}

Based on the results achieved in the previous section, here we assume the performance of a team is determined by the performance of its best player~\cite{dehpanah2021evaluating}.
Therefore, we take the maximum rating of team members to create the team's rating.

The calculated ratings can be used to predict the outcome of games~\cite{dehpanah2021player,shafiee2012study}.
In a field \textit{F}, denoting a head-to-head match (\textit{H2H}) between sides (either players or teams) $s_1$ and $s_2$, or a free-for-all match (\textit{F4A}) between sides $s_1$, $s_2$, ..., $s_n$, the prediction functions are defined as:

\begin{equation*}
\small
    \Phi_{H2H} =  \operatorname*{arg\,sort}_{i \in \{1,2\}} \Big((F,s_i)\: |\: \rho_{_{1,2}} \Big)
\end{equation*}

\begin{equation*}
    \Phi_{F4A} =  \operatorname*{arg\,sort}_{i \in \{1,2,...,n\}} \Big((F,s_i)\: |\: \rho_{_{1,2,...,n}}\Big)
    \small
\end{equation*}

\noindent where $\rho$ denotes the performance of teams and is represented by ratings created by rating systems as well as our behavioral single-factor ratings $\mu$, naive non-weighted hybrid ratings $\eta$, and weighted hybrid ratings $\Omega$.
The prediction function takes a list of the teams competing in a match along with their corresponding ratings and sorts teams based on their rating value in a descending order.
The ties are randomly broken by the models.
The resulted order is then returned by the functions as predicted ranks for that match.
Finally, the predicted ranks are compared with the observed ranks for each match to evaluate the performance of the models.


\section{Methodology}
\label{sec: methodology}
In this section, we first introduce the datasets used to perform our experiments.
We then describe some of the most important engineered behavioral features.
Finally, we explain our methodology and experimental setups in detail.

\subsection{Datasets}
Each dataset we used represents a certain game mode in the competitive shooter genre.

\subsubsection{Halo 3}
Halo 3 is one of the most played shooter games developed by Bungie Inc.
The game is the third installment in the popular Halo franchise and includes a wide variety of game modes and playing options such as Slayer and CTF. 
The Halo 3 dataset introduced by~\cite{delong2009project} consists of professional slayer and CTF matches between two teams of four players and is publicly available at \href{http://halofit.org/resources.php#datasets}{HaloFit}. 

\textbf{Slayer.}
Slayer (or Team Slayer) is a popular team \textit{deathmatch} in Halo 3 where two teams compete against each other with the goal of eliminating as many opponents as possible in a certain amount of time.
The slayer data includes in-game statistics such as number of kills, average time alive, and game outcomes for more than 2,300 matches and 270 unique players.

\textbf{CTF.}
\textit{CTF} refers to a popular objective-based game mode between two teams where players typically compete to steal the flag from the enemy's base and return it to their own base to earn points.
The CTF dataset includes in-game statistics such as number of kills, number of flags stolen, and game outcomes for more than 1,800 matches and 260 unique players. 

\subsubsection{CS:GO}
CS:GO is one of the most popular shooter games on Steam.
CS:GO's \textit{tactical head-to-head} game mode typically consists of two teams of five players, referred to as counter-terrorist and terrorist.
Terrorists attack certain locations on the map called bomb sites to plant a bomb and counter-terrorists attempt to defend bomb sites or defuse the bomb if it was successfully planted.
The CS:GO dataset is publicly available on \href{https://www.kaggle.com/mateusdmachado/csgo-professional-matches}{\textit{Kaggle}}.
After processing the raw data, the final dataset includes statistics such as number of kills, number of deaths, and game outcomes for around 20,000 matches and 4,500 unique players.

\subsubsection{PUBG: Battlegrounds (PUBG)}
PUBG is a popular \textit{battle royale} game developed and published by PUBG Corporation.
The dataset is publicly available on \href{https://www.kaggle.com/skihikingkevin/pubg-match-deaths}{\textit{Kaggle}}.
In this study, we considered PUBG duo matches (teams of two) to include small teams in our experimentation as well.
The dataset provides in-game statistics such as distance walked, number of kills, and rank for over 25,000 matches and 825,000 unique players.

\subsection{Engineered Behavioral Features}
In this section, we present some of the most important behavioral features engineered from the raw data.
The first group of created features mainly correspond to the player's skills, aggressiveness, and expertise.
These features are \textit{KD Ratio}, defined as the ratio between total number of kills to the total number of deaths, \textit{Killing Spree} defined as the average longest killing spree (eliminating more than one opponents without dying) achieved, \textit{Damage Dealt} defined as the average amount of damage dealt, \textit{Accuracy} defined as the average number of headshots registered, \textit{DBNO} (short for down but not out), defined as the average number of times player downed an opponent but did not finish them, \textit{Melee Kills}, defined as the average number of kills by melee attacks, and \textit{Grenade Kills}, defined as the average number of kills by grenade.

Another group of features inferred from the game outcomes include \textit{Winning Rate}, defined as the ratio between total number of wins to total number of games played, and \textit{Rank Ratio}, defined as the average rank percentage achieved, capturing a similar concept to winning rate for battle royale games.

The third group of features are more focused on player's interests and tactics.
These features include \textit{Survival}, defined as the average amount of time player stayed alive, \textit{Walking Distance}, defined as the average distance the player traveled on foot, and \textit{Riding Distance}, defined as the average distance the player traveled using a vehicle.

The fourth group of engineered behavioral features captures the support players provide for their teammates and how they help their teams in winning the game.
These features include \textit{Kill Assist}, defined as the average number of times player assisted teammates in eliminating an opponent, \textit{Flash Assist}, defined as the average number of times teammates eliminated an opponent blinded by player's flashbang, and \textit{Steal}, defined as the average number of times player stole the enemy's flag.

We also engineered behavioral features that negatively affect the player's and team's performance.
These features include \textit{Betrayal}, defined as the average number of times the player eliminated their own teammates, and \textit{Suicide}, defined as the average number of times the player eliminated themselves.

Finally, we considered the total number of games played as a behavioral feature that represents the player's \textit{Experience} and engagement with the game.

\subsection{Experimental Setups}

To engineer behavioral features and create behavioral ratings, we normalized all in-game statistics to eliminate scale differences.
For all datasets, we first sorted the matches by their timestamps.
We then retrieved the list of teams and players along with their corresponding ratings.
Players who appeared in the system for the first time were assigned default ratings.
The default ratings we considered for the rating systems are 1500 for Elo and Glicko, and 25 for TrueSkill.
We also initialized all engineered behavioral features (i.e., behavioral ratings) with zero.
For all rating approaches, we calculated the ratings of teams by taking the maximum of the ratings of team members.

We then sorted the teams based on their ratings and used the resulted order as the predicted ranks for the match.
For rating systems, players' ratings were updated after each match by comparing the predicted and observed ranks.
Behavioral ratings were also updated for all players after each match based on the events occurred within the match.
We evaluated the performance of all competing models using three setups hypothesizing that our behavioral ratings achieve more accurate rank predictions than rating systems for each setup.

The first setup considered all players in the system regardless of how skilled they are or how many games they played.
This setup includes players who played very few games.
Capturing the true performance level of these players is often impossible since there is no prior information available about them.
Therefore, rank prediction may be hampered for matches where teams consist of one or more of these players.

The next setup evaluated the predictive performance of competing models on the top-tier players in the system.
These players often win the games or achieve higher ranks and show consistent playing behavior.
We expect the models to achieve more accurate rank predictions for players with consistent behavior.
To identify these players, we sorted the players based on their most recent TrueSkill ratings and selected the top 50 players who had played more than 10 games.
Since these players had competed in different matches, we performed our evaluations on their first 10 games.

The last setup focused on the most frequent players in the data.
Playing more games often results in a more consistent playing behavior.
Therefore, similar to the top-tier players, we expect the competing models to achieve more accurate rank predictions for frequent players.
To identify the most frequent players, we selected all players who had played more than 100 games.
The predictive performance of competing models was then evaluated on their first 100 games.

Finally, we used Normalized Discounted Cumulative Gain (NDCG)~\cite{dehpanah2020evaluation,dehpanah2021evaluation,pourashraf2022using} and accuracy to evaluate rank predictions for battle royale and head-to-head games, respectively.


\section{Results and Discussions}
\label{sec: results}

In this section, we first present the results of the factor analysis we performed for detecting higher-level behavioral features. 
We then report classification results for two logistic regression models we used for determining feature weights.
Finally, we present the results of rank prediction using ratings from both rating systems and behavioral features.

\subsection{Factor Analysis Results}
We used principal component analysis for extracting factors and oblique method for rotation to allow a limited level of correlation among extracted factors.
The loadings were normalized to sum up to one for better interpretability.
Typically, with more diverse set of recorded in-game statistics comes higher chance of extracting different factors.
We identified one factor for Halo Slayer, one factor for Halo CTF, two factors for CS:GO, and two factors for PUBG.
We named these factors according to the nature of their building behavioral features.
Table~\ref{tab1} shows the extracted factors along with their building features and normalized loadings for each dataset.


\begin{table}[htbp]
  \centering
  \caption{Extracted factors along with their underlying features and normalized loadings}
  \begin{adjustbox}{width=180 pt,center} 
  \begin{tabular}{lllc}
    \hline
    \hline
    Datasets & Factors & Features & Normalized Loadings\\ 
    \hline
    \multirow{7}{*}{Halo-Slayer} & \multirow{7}{*}{Skill} & KD Ratio & 0.200283\\
     & & Killing Spree & 0.188712\\
     & & Survival & 0.178849\\
     & & Winning Rate & 0.171211\\
     & & Grenade Kills & 0.099063\\
     & & Accuracy & 0.082214\\
     & & Melee Kills & 0.079668\\
    \hline
     \multirow{5}{*}{Halo-CTF} & \multirow{5}{*}{Skill} & KD Ratio & 0.293236\\
     &  & Accuracy & 0.234795\\
     &  & Winning Rate & 0.192326\\
     &  & Grenade Kills & 0.142356\\
     &  & Kill Assist & 0.137287\\
    \hline
    \multirow{6}{*}{CS:GO} & \multirow{4}{*}{Skill} & Damage Dealt & 0.387052\\
     &  & KD Ratio & 0.270683\\
     &  & Accuracy & 0.183797\\
     &  & Winning Rate & 0.158467\\
    \cline{2-4} 
    & \multirow{2}{*}{Support} & Kill Assist & 0.669590\\
     &  & Flash Assist & 0.330410\\
    \hline
    \multirow{6}{*}{PUBG} & \multirow{3}{*}{Skill} & Damage Dealt & 0.344754\\
     &  & KD Ratio & 0.331903\\
     &  & DBNO & 0.323343\\
    \cline{2-4} 
    & \multirow{3}{*}{Strategy} & Survival & 0.396166\\
     &  & Walking Distance & 0.337053\\
     &  & Riding Distance & 0.266781\\
    \hline
    \hline

  \end{tabular}
  \end{adjustbox}

  \label{tab1}
\vspace{-2.5mm}
\end{table}


A common factor extracted in all datasets,`\textit{Skill}', is constructed from features that correspond to eliminating opponents and often correlate to the performance of top-tier players.
On the other hand, `\textit{Support}' refers to the situations where players help their teammates and corresponds to features denoting assists.
Finally, `\textit{Strategy}' represents features corresponding to tactics and playing style of players; a concept often connoting top-tier and experienced players.
As an example, the support $\mathcal{S}$ of player $p$ in CS:GO is estimated by:

\vspace{-2.mm}

\begin{equation*}
\small
\begin{split}
    \mathcal{S}(p) = 0.669590\big(Kill\:Assist(p)\big) + 0.330410\big(Flash \:Assist(p)\big)
\end{split}
\small
\end{equation*}

While the above factors were extracted from four separate datasets of completely different game modes, looking at their underlying features shows a high-level commonality specific to shooter games.
For example, regardless of game mode, a concept such as skill can be mainly inferred from components such as KD ratio or accuracy.
However, such components may defer depending on the available data.

\subsection{Classification Results and Feature Weights}
As mentioned earlier, we used two classification models for determining the contribution degree (i.e., weight) of behavioral features to players' performance; a binary logistic regression for our head-to-head games, and an ordinal logistic regression for our battle royale game.
We used a 5-fold cross-validation for evaluating the models and averaged their resulted scores across five folds.
Based on the results, our binary logistic regression models achieved 61.8\%, 63\%, and 65.4\% accuracy in the classification of ranks for Halo Slayer, Halo CTF, and CS:GO, respectively, and our ordinal logistic regression model reached a 72.2\% NDCG in classifying battle royale ranks.

As part of their output, the logistic regression models resulted in a list of the most important features in rank classification along with their weights.
We normalized the weights to sum up to one for better interpretability but we maintained their sign to capture the direction of features' effect on the overall performance.
The selected features and their normalized weights are shown in Table~\ref{tab2}.


\begin{table}[htbp]
  \centering
  \caption{Relevant features and their normalized weights}
  \begin{adjustbox}{width=170 pt,center}
  
  \begin{tabular}{l l c}
    \hline
    \hline
    Datasets & Relevant Features & Normalized Weights\\ 
    \hline
    \multirow{5}{*}{Halo-Slayer} & Skill & 0.330654\\
     & Experience & 0.320160\\
     & Kill Assist & 0.249425\\
     & Betrayal & -0.065018\\
     & Suicide & -0.034743\\
    \hline
    \multirow{6}{*}{Halo-CTF} & Skill & 0.330904\\
     & Steal & 0.249424\\
     & Experience & 0.219003\\
     & Betrayal & -0.076678\\
     & Melee Kills & 0.073562\\
     & Suicide & -0.050429\\
    \hline
    \multirow{3}{*}{CS:GO} & Skill & 0.552309\\
     & Experience & 0.276699\\
     & Support & 0.170992\\
    \hline
    \multirow{4}{*}{PUBG} & Strategy & 0.351819\\
     & Experience & 0.283292\\
     & Skill & 0.257153\\
     & Rank Ratio & 0.107736\\
    \hline
    \hline

  \end{tabular}
  \end{adjustbox}

  \label{tab2}
\vspace{-2.5mm}
\end{table}


Each engineered behavioral feature on its own constructs a rating referred to as single-factor ratings shown by $\mu$.
Our naive hybrid approach, denoted by $\eta$, includes a non-weighted summation of all single-factor ratings, regardless of their relevance or level of contribution to the performance of players and teams.
We used the most relevant features and their weights to create our third approach, a weighted hybrid denoted by $\Omega$.
For example, based on table~\ref{tab2}, the behavioral rating of player $p$ in CS:GO can be calculated as follows:
\vspace{-3mm}

\begin{equation*}
\small
\begin{split}
    \Omega(p) = 0.552309\big(Skill(p)\big) &+ 0.276699\big(Experience(p)\big) \\&+ 0.170992\big(Support(p)\big)
\end{split}
\small
\end{equation*}

The weights used in this approach can be periodically updated and optimized to ensure accurate representation of players.
For all three approaches, the rating of new players will be equal to zero since all features are initiated with zero.
Since our behavioral ratings are constructed from in-game statistics, lower rating values are associated with limited interactions and activities within games and vice versa.
For single-factor ratings, values may represent different meanings depending on the nature of the corresponding engineered feature.
For example, a lower rating in average distance walked may suggest a `camper' who may be an experienced player or an amateur player who just started playing the game and is not familiar with the map.
The meanings are often more clear when various features are considered together.
For example, the above representations can become more conclusive when we consider average distance walked with number of games played or KD ratio; features that more clearly represent experience and skills or aggressiveness.
While each single-factor rating may correlate differently with ranks, for our hybrid approaches, we assume players with higher ratings have more chance of winning a game.


\begin{table*}[htbp]

  \centering
  \caption{The predictive performance of rating systems and behavioral models-- \:$\mu_1$, $\mu_2$, and $\mu_3$: selected single-factor behavioral ratings, $\eta$: non-weighted hybrid, $\Omega$: weighted hybrid.}
\begin{adjustbox}{width=350 pt,center}  \begin{tabular}{l l | l l l| l l l | l l}
    \hline
    \hline
    Dataset & Setup & Elo & Glicko & TrueSkill & $\Phi(\mu_1)$ & $\Phi(\mu_2)$ & $\Phi(\mu_3)$ & $\Phi(\eta)$ & $\Phi(\Omega)$\\ 
    \hline
    
      \multirow{3}{*}{\vtop{\hbox{\strut Halo-Slayer}\hbox{\strut (\%Accuracy)}}} & All Players & $61.4\checkmark$ & 59.1 & 56.8 & 60.3 & 60.4 & 61.0 & 60.2 & $\mathbf{63.0^\star}$\\
     
    {} & Top-tier Players & 61.1 & 60.1 & $61.5\checkmark$ & \textbf{61.6} & \textbf{61.9} & \textbf{62.1} & 61.3 & $\mathbf{64.1^\star}$\\
     
    {} & Frequent Players & 59.8 & 61.3 & $61.8\checkmark$ & 61.2 & 61.3 & 61.5 & 60.7 & $\mathbf{63.2^\star}$\\
    \hline

     \multirow{3}{*}{\vtop{\hbox{\strut Halo-CTF}\hbox{\strut (\%Accuracy)}}} & All Players & $63.9\checkmark$ & 60.8 & 59.6 & 59.1 & 62.4 & \textbf{64.0} & 60.8 & $\mathbf{66.6^\star}$\\
     
    {} & Top-tier Players & 52.2 & 50.0 & $65.0\checkmark$ & 61.0 & 65.3 & \textbf{66.6} & 64.9 & $\mathbf{68.8^\star}$\\
     
     {} & Frequent Players & 56.3 & 59.2 & $62.8\checkmark$ & 60.7 & \textbf{63.1} & \textbf{63.1} & 61.0 & $\mathbf{67.7^\star}$\\
    \hline

     \multirow{3}{*}{\vtop{\hbox{\strut CS:GO}\hbox{\strut (\%Accuracy)}}} & All Players & $64.7\checkmark$ & 59.1 & 64.3 & 62.1 & 63.4 & 63.5 & 61.0 & $\mathbf{65.2^\star}$\\
     
     {} & Top-tier Players & 54.5 & $57.2\checkmark$ & 54.5 & \textbf{61.4} & \textbf{61.7} & \textbf{62.0} & \textbf{61.5} & $\mathbf{67.0^\star}$\\
     
     {} & Frequent Players & $64.3\checkmark$ & 59.2 & 63.6 & 61.2 & 62.0 & 62.4 & 62.2 & $\mathbf{66.1^\star}$\\
    \hline
    
     \multirow{3}{*}{\vtop{\hbox{\strut PUBG}\hbox{\strut (\%NDCG)}}} & All Players & 60.2 & 59.8 & $61.7\checkmark$ & \textbf{64.5} & \textbf{64.7} & \textbf{65.3} & 61.3 & $\mathbf{67.0^\star}$\\
     
     {} & Top-tier Players & $71.7 \checkmark$ & 69.6 & 69.2 & 70.3 & 71.6 & \textbf{72.4} & 70.1 & $\mathbf{79.2^ \star}$\\
     
     {} & Frequent Players & 66.7 & $67.9\checkmark$ & 60.3 & 66.2 & \textbf{69.3} & \textbf{69.7} & 67.6 & $\mathbf{76.7^\star}$\\
    \hline
    \hline

  \end{tabular}
  \end{adjustbox}

  \label{tab3}
\vspace{-2.5mm}
\end{table*}


\subsection{Rank Prediction Results}

For each match, we predicted the outcome using ratings calculated by our behavioral approaches as well as three popular rating systems; Elo, Glicko, and TrueSkill.
The prediction results are presented in table~\ref{tab3}.
The NDCG score in this table was calculated by taking the average NDCG values of all PUBG matches for each setup.
Each dataset includes several engineered behavioral features.
Therefore, for predictions using single-factor ratings, we included three features that resulted in the highest scores.

For each dataset and each setup, check marks show the highest score achieved among rating systems, bold numbers show the score of behavioral models outperforming rating systems, and stars denote the overall best performance.
The results of our experiments support our hypothesis for all datasets and setups showing that behavioral rating can be an effective progression-based substitute to ratings created by rating systems.
The superiority of our behavioral approach is specifically highlighted by the weighted hybrid approach.

In Halo Slayer dataset, while rating systems achieved a better performance than our single-factor models for the \textit{all players} and \textit{frequent players} setups, they were outperformed by all selected single-factor behavioral ratings (i.e., KD ratio, winning rate , and accuracy) for the \textit{top-tier players} setup.
As discussed before, these three features mainly represent skill or aggressiveness of players and their expertise in shooting; features that are often seen in highly skilled players in a fast-paced chaotic deathmatch game.   
The increase in prediction accuracy is even higher for our weighted hybrid approach compared to the best performing rating system in all setups.

In Halo CTF dataset, the rating systems were outperformed by winning rate in \textit{all players} and \textit{top-tier players} setups, and by winning rate and KD ratio in \textit{frequent players} setup.
While CTF games often allow for incorporating limited strategies, similar to slayer, behavioral features that imply skill, aggressiveness, and expertise provided a better representation of players here.
Again the best overall performance belongs to our weighted hybrid approach in all setups.

In CS:GO dataset, compared to our single-factor models, rating systems achieved a higher prediction accuracy for the \textit{all players} and \textit{frequent players} setups but they were outperformed by all selected single-factor behavioral ratings (i.e., damage dealt, winning rate, and KD ratio) for the \textit{top-tier players} setup.
For CS:GO, features that directly demonstrate the skill and expertise of players provided a better representation of players' behavior compared to features that imply supporting play styles.
The best performance, again, belongs to our weighted hybrid approach in all setups.

Finally, in PUBG dataset, our single-factor ratings (i.e., riding distance, KD ratio, and experience) outperformed the rating systems in \textit{all players} setup.
On the other hand, the only single-factor rating that could outperform rating systems in \textit{top-tier players} setup was KD ratio.
For the \textit{frequent players} setup, survival and experience outperformed the best performing rating system.
As oppose to the other three datasets, in PUBG, features that imply players' strategies and their engagement or experience with the game provided more accurate representations of their behavior compared to features that demonstrate their skills.
This is in part due to the availability of more diverse in-game statistics for PUBG that enabled capturing behavioral aspects beyond skills.
The PUBG dataset also involved many more matches compared to other datasets that enabled our progression-based approaches to yield more consistent representations of behavioral features by observing more games over time.
Similar to the previous three datasets, the best overall performance belongs to our weighted hybrid approach in all setups.

A crucial aspect of gameplay that was missing in all our datasets is movement dynamics.
Features demonstrating how players walk or sprint within the game, how they move when they are shooting an opponent, or whether they perform certain movements could make a clear distinction between different groups of players.
For example, top-tier players use movements like bunny hopping, tap strafing, or crouch spamming to avoid taking damage in a fight while amateur players often stand still when shooting the enemy making them an easy target to the opponents.
The behavioral rating approach could highly benefit from a rich source of in-game statistics; statistics that capture various interactions of players within the game.

Nonetheless, our created ratings benefit from an innate interpretability because they do not include unknown features similar to those created by latent factor models.
Instead, our proposed ratings consist of prevalent behavioral features well-known to players.
In addition, unlike both traditional rating systems and their extensions, our created ratings do not solely rely on the game outcomes or few aspects of the gameplay to model players.
Instead, they consider different aspects of the playing behavior of players as well as a wide variety of their interactions within the games they played to generate more accurate representations of their expected performance.


\section{Conclusion and Future Works}
\label{sec: conclusion}

In this paper, we presented a novel approach for creating accurate and interpretable representation of players' behavior in competitive online shooter games.
To address different aspects of a player's behavior, we performed our experiments on four datasets each of which representing a unique game mode where players demonstrate different behaviors based on the game's objective and rules.
We created several behavioral ratings that represented our player models.
To test the accuracy of the behavioral ratings, we used the created ratings to predict the players' performance.
We then compared the predictive performance of our behavioral ratings with that of three popular rating systems.
The experimental results suggested the superiority of behavioral ratings over representations created by rating systems in estimating the true performance level of players, particularly when the behavioral ratings are incorporated into a weighted hybrid.

As future work, we plan to incorporate created behavioral features into the rating systems to reinforce their estimation power.
We also plan to use the created player models to enhance the quality of matchmaking by matching players and teams based on different aspects of their playing behavior.

\bibliographystyle{IEEEtran}
\bibliography{IEEEabrv, reference}
\end{document}